# Mobility Prediction in Wireless Ad Hoc Networks using Neural Networks

Heni KAANICHE and Farouk KAMOUN

**Abstract**—Mobility prediction allows estimating the stability of paths in a mobile wireless Ad Hoc networks. Identifying stable paths helps to improve routing by reducing the overhead and the number of connection interruptions. In this paper, we introduce a neural network based method for mobility prediction in Ad Hoc networks. This method consists of a multi-layer and recurrent neural network using back propagation through time algorithm for training.

——————————— ◆ ———————————

## 1 INTRODUCTION

MOBILE Ad hoc networks (MANETs) [1] are composed of mobile nodes connected by wireless links. All nodes can freely and dynamically self-organize into arbitrary and temporary "Ad Hoc" network topologies. MANETs are self-organizing networks because they do not use any infrastructure such as base station or router. This implies that every node performs as a host as well as a router since it is in charge of routing information between its neighbors, contributing to and maintaining connectivity of the network. Thus, in a MANET, the routing protocol used is of primary importance because it determines how a data packet is transmitted over multiple hops from a source node to a destination node.

The route formation should be performed rapidly, with minimal overhead. The routing protocol must also adapt to frequently changing network topologies caused by nodes mobility, as well as other network characteristics. Various routing schemes have been proposed for MANETs [2-6]. The majority of these solutions are based on finding the shortest path in term of distance or delay. However, the shortest path may break up quickly after its establishment. Indeed, because of nodes mobility, some of links on the shortest path may fail as soon as the path is established. This failure causes connection interruption and data loss, if the rediscovering routes phase is not accomplished rapidly. However, rediscovering routes phase involves a substantial overhead. So, routing based on selecting the shortest path leads to degradation in the routing quality of service. This is why stable paths are worth to be exploited for routing packets, instead of shortest paths [7-8]. The paths stability estimation can be done by predicting nodes future locations, which we call mobility prediction. In this study, we propose a method for mobility prediction in Ad Hoc networks. This method is based on a recurrent neural network.

————————————————

• Heni.KAANICHE is with the faculty of Sciences of Sfax, Tunisia.
• Farouk.KAMOUN is with Computer Science School of the University of Manouba, Tunisia.

This paper is organized as follows: In section 2, we give a definition of mobility prediction. In section3, we justify the importance of mobility prediction in routing. We also calculate path expiration time which quantifies the path stability. In section 4 we propose a technique for mobility prediction, which is a multi-layer and recurrent neural network using back propagation through time algorithm for training. Section 5 illustrates the effectiveness of the proposed mobility predictor. Finally, some concluding remarks are presented in Section 6.

## 2 DEFINITION OF MOBILITY PREDICTION

Mobility prediction of a node is the estimation of their future locations. The definition of "location" depends on the kind of wireless network: In infrastructure networks, location means the access point to which the mobile terminal is connected. Many location prediction methods are proposed in literature: [13]. The main advantage of location prediction is to allocate, in advance, the convenient next access point before the mobile terminal leaves its current one, in order to reduce the interruption time in communication between terminal mobiles. In without infrastructure networks or MANETs, mobile's location means its geographic coordinates. Location prediction in Ad Hoc networks is a new topic. Its main advantage is to estimate link expiration time [7],[9] in order to improve routing performances. Authors of [7] and [9] proposed two different methods for mobility prediction. However, these methods assume that nodes move according the RWM (Random Waypoint Mobility) model [10]. As a result, nodes mobility prediction moving according to other models can lose its accuracy and efficiency.

In this study, we interest in developing a technique for mobility prediction in Ad Hoc networks. This technique is independent of the mobility model used by nodes. We assume that each node in the Ad Hoc network is aware of its location. Most commonly, the node will be able to learn its location using an on board GPS (Global Positioning System) receiver. So, it can periodically record its geographical location. All the locations recorded, define the node trajectory. At time tk., this trajectory can be noted by: (Xtk-i , Ytk-i,, Ztk-i){i=0..p}, where p is the number of locations recorded before the current time tk (see Fig. 1). Knowing its previous locations (Xtk-i , Ytk-i,, Ztk-i){i=0..p}, a mobile node can estimate their future ones (Xtk+i , Ytk-i,, Ztk+i){i=0..N}, where N is the number of predicted locations.

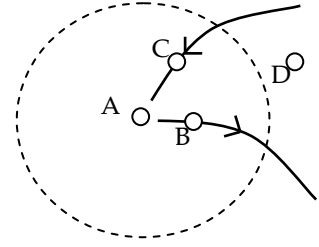

Fig. 2. Simple mobility scenario in an Ad Hoc network.

### 3.1. Path Expiration Time

The path stability is quantified by the Path Expiration Time (PET), given by the expression 1.

$$LEP_S(D) = \min_{k=1:k_0}(LET_{N_k}(N_{k-1})) \qquad (1)$$

where, $LEP_S(D)$ is the expiration time of the path joining S and D; $LET_{N_k}(N_{k-1})$ is the expiration time of the link connecting the node $N_k$ and the node $N_{k-1}$; $k$ is an integer where $1 < k \leq k_0$ ; $N_{k0}$ = D and $N_1$ = S.

### 3.2. Link Expiration Time

The determination of the LET is illustrated by the following example given by Fig. 3. N1 and N2 are two mobile nodes. At the moment tk, N1 wishes to calculate the expiration time of the link connecting it to N2 (LETN1(N2)). At this moment, geographical coordinates of N1 and N 2 are N1(X1tk , Y1tk, Z1tk) and N2(X2tk , Y2tk, Z2tk), respectively. Using a technique of mobility prediction based on previous locations, both N1 and N2 estimate their N future locations: (X1tk+i , Y1tk+I, Z1tk+i){i=0 to N} and (X2tk+i , Y2tk+I, Z2tk+i){i=0 to N}, respectively. At first, N1 should dispose of its future locations as well as those of N2. Then, it computes the distances dtk+i{i=0 to N}, separating it to N2 at times tk+i {i=0 to N}. The LET corresponds to a distance separating N1 and N2 equals to 250m, which represents the transmission range of N1.

The computation of LET is detailed as follow: We consider the function that associates the distance d between N1 and N2 to time t. This function can be approximated by a N-1 degree polynomial P(t). The coefficients ai of P(t) are determined by the resolution of the following system of equations:

$$\begin{cases} \sum_{i=0}^{N-1} a_i t_{k+1}^{N-1-i} = d_{k+1} \\ \sum_{i=0}^{N-1} a_i t_{k+2}^{N-1-i} = d_{k+2} \\ \quad\vdots \\ \sum_{i=0}^{N-1} a_i t_{k+N}^{N-1-i} = d_{k+N} \end{cases} \qquad (2)$$

Finally LET is calculated by resolving the following equation:

$$P(LET) = 250 \qquad (3)$$

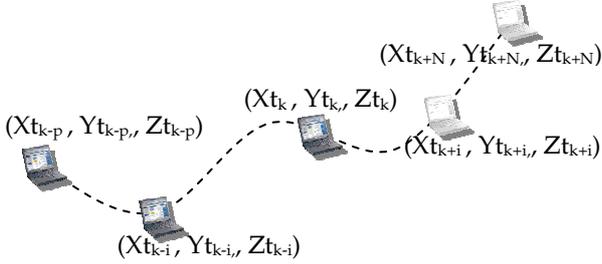

(Xt_{k-p}, Yt_{k-p}, Zt_{k-p})
(Xt_{k-i}, Yt_{k-i}, Zt_{k-i})
(Xt_k, Yt_k, Zt_k)
(Xt_{k+i}, Yt_{k+i}, Zt_{k+i})
(Xt_{k+N}, Yt_{k+N}, Zt_{k+N})

Fig. 1. Mobility prediction of a mobile node

Time series is a set of observations from past until present. Time series prediction is to estimate future observations. Location prediction is a particular case of time series prediction. Indeed, the mobile node's trajectory defined by locations (Xtk-i , Ytk-i,, Ztk-i){i=0..P} represents three time series: (Xtk-i){i=0..P}, (Ytk-i){i=0..P} and (Ztk-i){i=0..P}. So, location prediction problem can be resolved by the prediction of these three time series to obtain (Xtk+i){i=0..N}, (Ytk+i){i=0 ..N} and (Ztk+i){i=0..N}. This study aims for developing a technique for long-term time series prediction to be exploited for mobility prediction.

## 3. MOBILITY PREDICTION AND ROUTING IN AD HOC NETWORKS

This section shows the importance of mobility prediction in routing Ad hoc networks, by giving a simple mobility scenario. Fig. 2 represents an Ad Hoc networks containing four nodes which are A, B, C and D. A is stable, C moves slowly towards A and B moves rapidly away from A and D. A has data packets to send to D. It finds that to reach D, packets can pass either through B, or through C. If A chooses B as intermediate node, then the communication will not last long time since the link (A,B) will be rapidly broken, due to the mobility of B. But if A, takes into account the mobility of B and C, it will choose C as intermediate node because the expiration time of the link (A,C) is superior to that of (A, B), since C have chance to remain in A transmission range, more than B. The fact that A chooses C as next hop to reach D contributes to the selection of the path which has the greatest expiration time or the most stable path. According to this example, it is clear that selecting the most stable path can avoid future link failure, which improves routing.

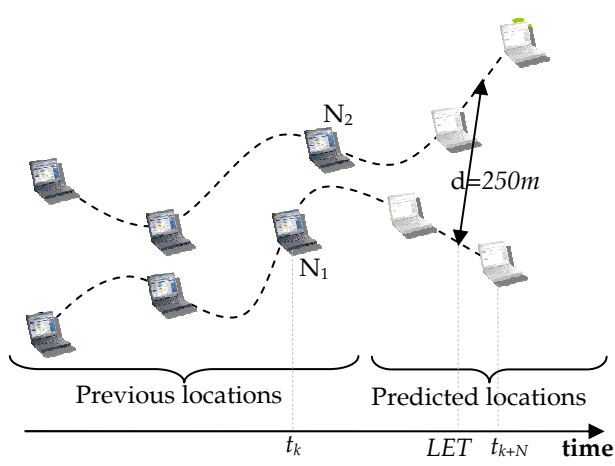

Fig. 3. Determination of the link expiration time

In conclusion, mobility prediction can improve considerably routing by selecting stable route and consequently, avoiding route failure and substantial overhead. In the following section, we present a time series prediction technique based on recurrent neural network and providing long-term prediction (multi-step prediction). This technique can be applied to a time series. In particular, it can be applied to location time series or mobility prediction.

## 4. RECURRENT NEURAL NETWORK BASED PREDICTION

According to literature, time series prediction techniques can be classified in two categories: statistical techniques [14] and techniques based on artificial intelligence tools such as neural networks, fuzzy systems and neuro-fuzzy systems [15]. In particular, recurrent neural network, showed their efficiency for multi-step time series prediction [15], [18], [19], [22]. Indeed, recurrent neural networks include internal dynamic memory, created from its structure containing cycles. This feature grants a dynamic mapping relating input to output sequences. Recurrent neural networks propose several architectures and training algorithms. The efficiency of the neural prediction technique depends on the choice of the architecture and the training algorithm.

### 4.1. Recurrent neural Network architecture for prediction

Neural architecture considered in this investigation is a recurrent and multi-layer neural network (see Fig. 4). This architecture is composed of three layers arranged in a feedforward fashion: The first layer is the input layer which represents the dynamic memory of network. This memory is originated by a feedback between the output layer and the input layer; as well as feedbacks between neurons themselves from input layer. The network input layer can contain external inputs coming from time series observations, as well as estimated input coming from previous network output. In fact, this architecture is used N times (N-step prediction) to estimate N observations of time series. In each prediction step $k(k \neq N)$, network output $s(t + k)$ is stored in the input layer to be able to estimate the following value $s(t + (k + 1))$, that corresponds to prediction step (k+1). This storage is due to the feedback connection between the output neuron and a neuron in the input layer.

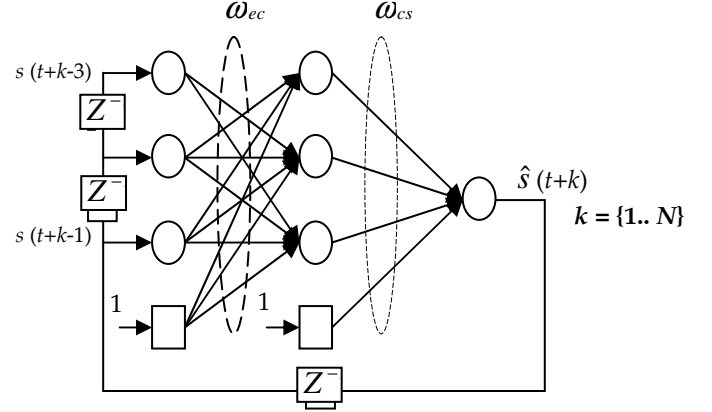

Fig. 4. Neural architecture for the prediction

The second layer is called hidden layer. The role of this layer is to store the characteristics of the learned patterns, to increase the freedom degree in the problem resolution and to improve generalization ability of the network, just like the traditional multi-layer.

### 4.2. Notation

To cope with the structural complexity of neural networks, the following notational convention is employed. The input, hidden and output layers contain Ne, Nc and one neurons, respectively. Let $\omega_{ec}$ denotes the weight of the link connecting neuron e in the input layer and neuron c in the hidden layer and $\omega_{cs}$ denotes the weight of the link connecting neuron c in the hidden layer and neuron s in the output layer, e ={1..Ne}, c ={1..Nc} and s =1. Bias terms are included, by adding a bias neuron in both input and hidden layers, whose input have constant value equal to one (see Fig. 4).

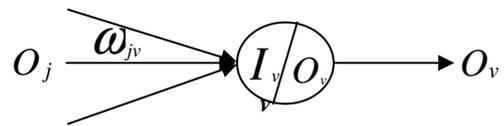

Fig. 5. Notation of the neuron's input/output

Let v denotes a neuron belonging to one of treatment layer (hidden and output i.e. v = c or s). Each neuron v computes a weighted sum of the inputs denoted by $I_v$.

The quantity $I_v$ computed at each neuron becomes then argument of a sigmoid activation function, which resides in the neuron itself. Hence, the value returned by the activation function of neuron v is $O_v$ (see Fig. 5 and Equation 4).

$$I_v = \sum_j \omega_{jv}.O_J, \ O_v = f(I_v),$$

$$f: x \longmapsto \frac{1}{1+e^{-x}}, f'(x)=f(x).(1-f(x)) \quad (4)$$

## 4. 3. Learning algorithm

In prediction step k, real observation r(t+k) is forecasted as $\hat{s}$(t+k), that correspond to the network output. The quantity $J_k$, represents the square error function in prediction step k:

$$J_k = \frac{1}{2}\left[\hat{s}(t+k) - r(t+k)\right]^2 \quad (5)$$

The total error function in prediction step N is the sum of the error functions $J_k$ that corresponds to prediction step k:

$$J = \sum_{k=1}^{N} J_k = \sum_{k=1}^{N} \frac{1}{2}\left[\hat{s}(t+k) - r(t+k)\right]^2 \quad (6)$$

The network is trained using BPTT (Back Propagation Through Time) algorithm by minimizing the error function J. The weights are adapted at the time (t + N) which corresponds to the  N th prediction step, using the function error J

$$\Delta \omega_{uv} = -\varepsilon \ \frac{\partial J}{\partial \omega_{uv}}$$

$$= -\varepsilon \sum_{k=1}^{N} \frac{\partial J_k}{\partial \omega_{uv}}$$

$$= -\varepsilon \sum_{k=1}^{N} \frac{d J_k}{d\hat{s}(t+k)} \cdot \frac{d\hat{s}(t+k)}{d\omega_{uv}}$$

$$= -\varepsilon \sum_{k=1}^{N} \left(\hat{s}(t+k) - r(t+k)\right).$$

$$\left(\sum_{j=1}^{n} \frac{\partial \hat{s}(t+k)}{\partial \hat{s}(t+k-j)} \cdot \frac{d\hat{s}(t+k-j)}{d\omega_{uv}} + \frac{\partial \hat{s}(t+k)}{\partial \omega_{uv}}\right) \quad (7)$$

n denotes the number of forecasted inputs, (because the network output is reinjected in the input layer).

The quantities $\frac{\partial \hat{s}(t+k)}{\partial \hat{s}(t+k-j)}$ and $\frac{\partial \hat{s}(t+k)}{\partial \omega_{uv}}$ are given

by:

$$\frac{\partial \hat{s}(t+k)}{\partial \hat{s}(t+k-j)}$$

$$= \sum_{c=1}^{Nc} \frac{\partial \hat{s}(t+k)}{\partial I_s} \cdot \frac{\partial I_s}{\partial O_c} \cdot \frac{\partial O_c}{\partial I_c} \cdot \frac{\partial I_c}{\partial \hat{s}(t+k-j)}$$

$$=f'(I_s). \ \sum_{c=1}^{Nc} \omega_{cs}. f'(I_c).\omega_{ec} \quad (8)$$

The weight $\omega_{ec}$ denotes the weight of the link connecting the neuron e in the input layer, whose input is equal to s(t+k-j) and a neuron c in the hidden layer.

$$\frac{\partial \hat{s}(t+k)}{\partial \omega_{uv}} = \frac{\partial \hat{s}(t+k)}{\partial I_v} \cdot \frac{\partial I_v}{d\omega_{uv}} \quad (9)$$

Two cases must be considered:

First case: $\omega_{uv} = \omega_{ec}$ (i.e. u belongs to input layer and v belongs to hidden layer)

$$\frac{\partial \hat{s}(t+k)}{\partial \omega_{ec}} = \frac{\partial \hat{s}(t+k)}{\partial I_c} \cdot \frac{\partial I_c}{d\omega_{ec}}$$

$$= \frac{\partial \hat{s}(t+k)}{\partial I_s} \cdot \frac{\partial I_s}{\partial O_c} \cdot \frac{\partial O_c}{\partial I_c} \cdot \frac{\partial I_c}{d\omega_{ec}}$$

$$= f'(I_s) \ \omega_{cs}. f'(I_c). O_e \quad (10)$$

Second case: $\omega_{uv} = \omega_{cs}$ (i.e. u belongs to hidden layer and v belongs to output layer)

$$\frac{\partial \hat{s}(t+k)}{\partial \omega_{cs}} = \frac{\partial \hat{s}(t+k)}{\partial I_s} \cdot \frac{\partial I_s}{d\omega_{cs}}$$

$$= f'(I_s). O_c \quad (11)$$

From equations (6) - (11), equation (7) is given by:

$$\frac{\partial J}{\partial \omega_{ec}} = \sum_{k=1}^{N} \left(\hat{s}(t+k) - r(t+k)\right)^{\bullet}$$

$$\left\{\sum_{j=1}^{n} [f'(I_s).\sum_{c=1}^{Nc}(\omega_{cs}. f'(I_c).\omega_{jc}).g_{k-j}] + g_k\right\} \quad (12)$$

where :

- $g_k = f'(I_s) \ \omega_{cs} f'(I_c). O_e$ , when the output is $\hat{s}(t+k)$ i.e. $f'(I_s) = \hat{s}(t+k-j).[1-\hat{s}(t+k-j)]$

- $g_{k-j} = f'(I_s) \ \omega_{cs} f'(I_c). O_e$ , when the output is $\hat{s}(t+(k-j))$, i.e. $f'(I_s) = \hat{s}(t+k).[1-\hat{s}(t+k)]$

$$\frac{\partial J}{\partial \omega_{cs}} = \sum_{k=1}^{N} \left(\hat{s}(t+k) - r(t+k)\right)^{\bullet}$$

$$\left\{\sum_{j=1}^{n} [f'(I_s).\sum_{c=1}^{Nc}(\omega_{cs}. f'(I_c).\omega_{jc}).g_{k-j}] + g_k\right\}$$

(13)

Where:

- $g_k = f'$ when the output is $\hat{s}(t+k)$

- $g_{k-j} = f'(I_s). O_c$ , when the output is $\hat{s}(t+(k-j))$

# 5. Test of the neural predictor

In this section, we test the neural predictor on time series describing locations of an Ad hoc mobile node, in order to test the efficiency of the predictor in mobility prediction.

The movement of an Ad hoc mobile node can be described by three location time series: x(t-i){i=0..P }, y(t-i){i=0..P} and z(t-i){i=0..P}. The particularity of such a series is that it follows a specific mobility model. In fact, to simulate the movement of an Ad hoc mobile node, the literature offers a set of mobility models [10]: RWM (Random Walk Mobility), RWM (Random Waypoint Mobility), ABSAM (A Boundless Simulation Area Mobility), GM (Gauss-Markov), CSM (City Section Mobility).

The majority of the searchers use RWM model in their Ad hoc networks simulations because it is flexible and simulates in a realistic way the movement of people [10]. So, we chosen RWM model to construct location time series. In this model, a mobile node starts from its current position and chooses randomly a destination position in its mobility territory. Then, it moves straight towards this destination with a constant speed, chosen randomly in an interval of speeds. When it reaches the destination, it remains stable during a period of time then restarts the same movement process.

## 5.1. Selection of recurrent neural network structure

The selection of the numbers of input neurons (Ne) and hidden layer neurons (Nc), can be done by heuristic methods. The method that we adopted is as follows: We consider thirty location time series based on RWM model, containing 400 patterns each one: the first 200 patterns are used for training, the rest of patterns for the test (generalization). Each series is applied to the network while varying Ne and Nc in the interval [1..30]. For each combination of the couple (Ne, Nc), we calculate the training error and the generalization error (formula 15). The following observations were then retained:

- Not like Ne, a small variation of Nc can affect the prediction accuracy.

- A great value of Ne can lead to an underlearning and a very small value of Ne can lead to an overfitting.

- A very good training, i.e. training with a very small error can affect the generalization ability of the neural network, this is why, we choose the generalization error as criterion in the selection of the Ne and Nc parameters and not the training error.

- The number Nc depends on the length of subpatterns provided for training, i.e.Ne, and the number Ne depends only on the training data [22]. So we fixed Nc to determine optimal value of Ne, i.e. that provides the smallest generalization error. We found that 70% of location time series have optimal Ne between 5 and 10. For the rest of series (30 %), we note that the generalization error obtained by theirs optimal Ne is not far (about 0.01) from that of the firsts series. Thus, we choose Ne equal to 8, being sure that we will not penalize some particular series.

- The variation of Nc doesn't affect greatly the prediction accuracy, starting from Nc equal to 5. The choice of an elevated value of Nc will increase the number of parameters to be estimated and training delay, without improving considerably the generalization ability. This is why, we choose Nc equal to 5.

$$E_{train/gener} = \sum_{sam-} \sum_{k=1}^{3} \frac{1}{2} \left[ \hat{s}(t+k) - s(t+k) \right]^2 \qquad (15)$$

## 5.2. Testing the selected predictor architecture

Let us consider an Ad hoc mobile node which moves according to RWM mobility model with a speed varying in [0..20]. Its coordinated are recorded each 10s, starting from 0s (initial time), until 4000s. So we obtain two location time series x(t-i) {i= 0. 400} and y(t-i) {i= 0. 400}. Now, we will test the predictor on his two location time series inorder to forecast the mobile node movement (trajectory). The first 200 co-ordinates are used for training and the rest for generalization. Training as well as generalization (test) was done on a horizon of three steps.

The parameters Ne and Nc are fixed respectively at 8 and 5. Fig. 6 illustrates the test of the neuronal predictor on the two series x and y. The forecasted trajectory of the mobile is deduced from predicting x and y (Fig. 7).

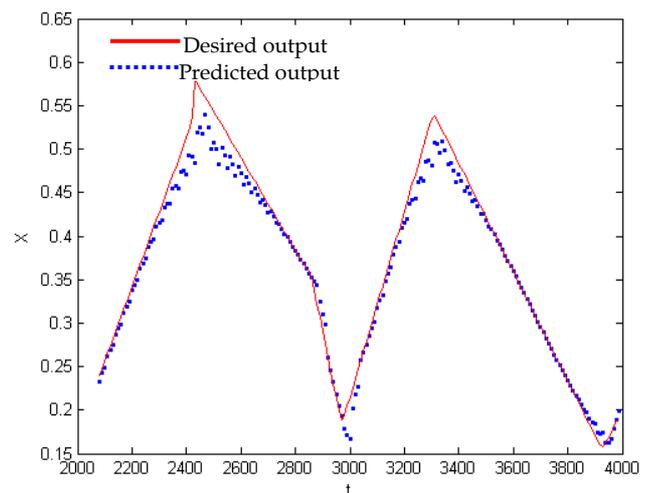

Fig. 6.a. Test of time series x

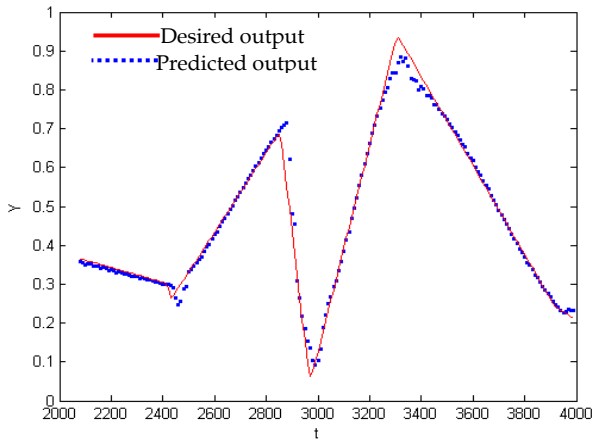

Fig. 6.b. Test of time series y

Fig. 6. Prediction of tow location time series

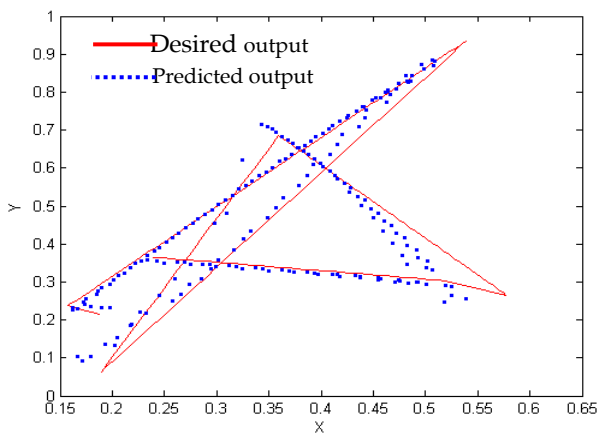

Fig. 7. Prediction of the mobile trajectory

## 7. Conclusion

Mobility prediction in Ad Hoc networks is a very important matter because it can improve significantly routing in wireless Ad Hoc networks, by reducing the overhead and the number of connection interruptions. In this study, a recurrent neural network has been proposed for long-term time series prediction, since mobility prediction is a particular problem of time series prediction. The architecture of this neural predictor is a three-layer network with feedback connections. Back propagation through Time algorithm has been used to train the recurrent neural network. To exam the efficiency of the predictor in mobility prediction, we have tested the neural predictor on time series describing locations of an Ad hoc mobile node, moving according to RWM model.

*Heni KAANICHE* received the Engineering Degree from the National School of Engineering of Sfax, Tunisia, in July 2001 and he completed his master in the Network and Performance team of the LIP6 laboratory. He received his Master Degree in Computer Networks and Telecommunications from the University Pierre and Marie Curie (Paris VI), France, in September 2002. He is currently a Ph.D student under the direction of Professor Farouk KAMOUN in the Network team of the CRISTAL laboratory at the National School of Computer Science, ENSI, Manouba, Tunisia. His main research concerns Wi-Fi, Mobility and Ad Hoc networks.

*Farouk KAMOUN*, PhD. (born 1946) is a Tunisian computer scientist and professor of computer science at ENSI (Ecole Nationale Sciences Informatiques : Computer Science School of The University of Manouba, Tunisia). He contributed in the late 1970s to significant research in the field of computer networking in relation with the first ARPANET network. He is also one of the pionniers of the development of the Internet in Tunisia in the early 1990s.